# Design and Implementation of a Kalman Filter-Infused Algorithm for Tilt Estimation

Yuehan Ma [1]
*Independent Researcher*
yuehanma.com
[1]lucasmahjkl@gmail.com

Hongji Dai [2]
*Independent Researcher*
[2]terback@gmail.com

## Abstract

Accurate tilt angle estimation is important in many engineering applications, such as robotics, motion tracking, and embedded control systems. However, measurements from low-cost inertial sensors are often degraded by noise and drift. This paper presents a single-axis tilt angle estimation system based on the MPU6050 inertial measurement unit, implemented on an RP2040 microcontroller platform, with sensor fusion achieved through a Kalman filter.

The accelerometer provides a direct estimate of tilt angle from gravity but is sensitive to noise and short-term fluctuations. The gyroscope provides smooth angular rate measurements, but integration over time introduces drift. To overcome these limitations, a Kalman filter is used to combine measurements from both sensors, leveraging the long-term stability of the accelerometer and the short-term smoothness of the gyroscope.

Both simulation and hardware experiments are performed. In simulation, sensor noise and drift are modeled to evaluate the filter performance under control conditions. In the hardware implementation, real-time MPU6050 data is acquired and processed by the RP2040 platform, and the estimated tilt angle is compared with accelerometer-only and gyroscope-only outputs. The results show that the proposed method effectively reduces noise measurements and suppresses long-term drift while preserving good dynamic response.

Overall, the system provides more stable and accurate tilt estimation than either sensor alone, demonstrating a practical and accessible approach for Kalman filter based sensor fusion in embedded application. This manuscript is a preprint version of the work.

## 1. Introduction

Accurate tilt angle estimation plays a critical role in a wide range of engineering applications, including robotics, motion tracking, and embedded control systems [3][9]. In these systems, reliable orientation information is essential for tasks such as stabilization, navigation, and human–machine interaction. With the increasing availability of low-cost inertial measurement units (IMUs), it has become feasible to implement tilt estimation in compact and affordable embedded platforms. However, achieving high accuracy using such sensors remains a significant challenge [2][5].

Low-cost IMUs, such as the MPU6050 sensor module, typically combine an accelerometer and a gyroscope to provide motion and orientation-related measurements [7]. Each of these sensing modalities has inherent limitations. The accelerometer can estimate tilt angle based on the direction of gravity, but it is highly sensitive to noise and external vibrations, resulting in significant short-term fluctuations [2][5]. On the other hand, the gyroscope provides smooth angular velocity measurements, but estimating angle through time integration introduces cumulative errors, leading to long-term drift [2][5]. As a result, using either sensor independently does not provide sufficiently accurate or stable tilt estimation.

To overcome these limitations, sensor fusion techniques are commonly employed to combine the complementary characteristics of accelerometers and gyroscopes [4][8]. Among these techniques, the Kalman Filter is widely used due to its ability to optimally estimate system states in the presence of noise and uncertainty [1][3]. By integrating prediction from the gyroscope with correction from the accelerometer, the Kalman Filter can achieve a balance between short-term smoothness and long-term stability, making it particularly suitable for embedded tilt estimation systems [3][5].

In this work, a tilt angle estimation system based on the MPU6050 IMU is developed and implemented on an RP2040-based microcontroller platform, specifically the STEPico RP2040 development board. A real-time Kalman Filter is designed to fuse accelerometer and gyroscope measurements, enabling improved accuracy and stability in tilt estimation. The system is designed with a focus on accessibility and practicality, making it suitable for educational and low-cost embedded applications.

This work makes the following contributions:

- Design and implementation of a low-cost tilt angle estimation system using the MPU6050 IMU on an RP2040 microcontroller platform.
- Development of a real-time Kalman Filter-based sensor fusion algorithm for embedded systems.
- Comprehensive quantitative evaluation of the system, including noise analysis, drift analysis, and accuracy assessment under controlled conditions.
- A qualitative user-level evaluation demonstrating the impact of improved estimation on interaction smoothness in an embedded display application.

## 2. System Design

### 2.1 Hardware Architecture

The tilt estimation system is built using a low-cost inertial sensing module, the MPU6050 sensor module, and an RP2040-based microcontroller platform, specifically the STEPico RP2040 development board. The MPU6050 integrates a 3-axis accelerometer and a 3-axis gyroscope, providing raw motion and orientation-related data required for tilt estimation. This study will focus on single-axis tilt estimation to simplify sensor fusion complexity, effectively evaluating the noise suppression of the Kalman filter algorithm in the accelerometer and gyroscope.

The MPU6050 communicates with the microcontroller via the Inter-Integrated Circuit (I2C) interface. The microcontroller periodically acquires accelerometer and gyroscope measurements at a fixed sampling rate. In this implementation, a sampling frequency of 100 Hz is used to balance computational load and estimation accuracy.

The sensor module is mounted on a rigid platform to ensure consistent orientation during experiments. This reduces unintended disturbances and improves measurement repeatability. The microcontroller processes the incoming sensor data in real time and outputs the estimated tilt angle through a serial interface for logging and analysis. In addition, the system can optionally interface with a small display module for real-time visualization in interactive applications. As illustrated in Figure 1, the MPU6050 provides acceleration and rotation data over I²C to the RP2040, which converts and filters the data using the Kalman algorithm continuously to output accurate angle measurements.

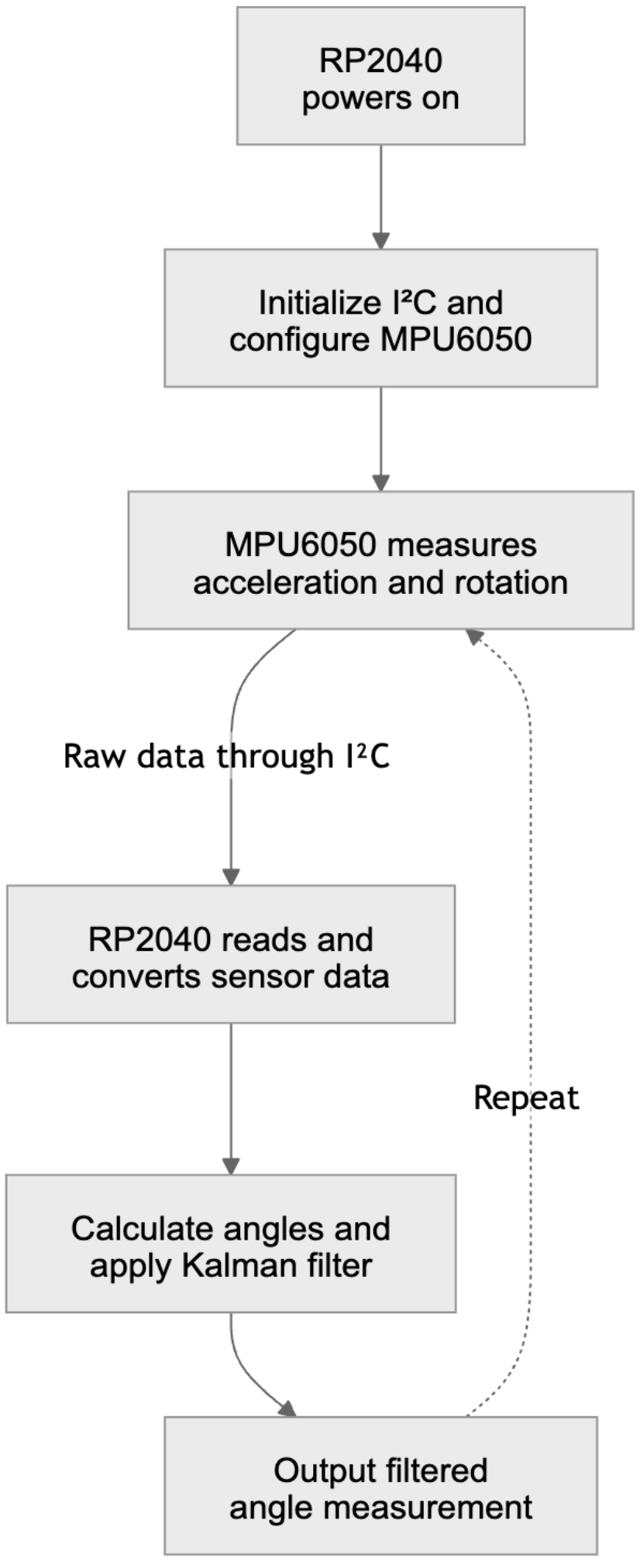

*Fig 1: Data acquisition and angle-estimation workflow between the RP2040 STEPico microcontroller and MPU6050 sensor.*

### 2.2 Software Architecture

The software system is organized as a real-time processing pipeline that converts raw sensor measurements into a stable tilt angle estimate. The overall data flow is shown conceptually as:

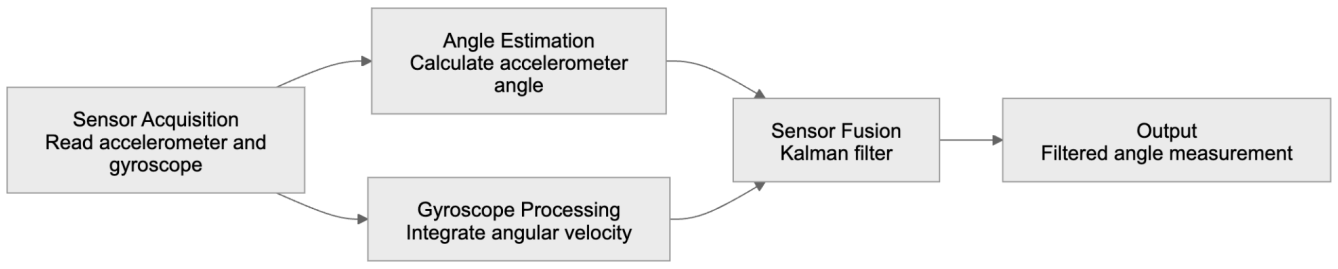

*Figure 2. Block diagram of the Kalman filter–based angle estimation process*

At each sampling step, raw accelerometer and gyroscope data are first read from the MPU6050. These measurements are then processed to obtain two intermediate angle estimates:

- An accelerometer-based angle derived from the gravity vector
- A gyroscope-based angle obtained through angular rate integration

These intermediate estimates are then fused using a Kalman Filter, which produces the final tilt angle output. The resulting angle is transmitted via serial communication for data logging and can also be used to drive real-time applications such as display-based interaction.

### 2.3 Data Processing Pipeline

The processing pipeline consists of four main stages:

*1. Sensor Data Acquisition*

Raw data from the accelerometer and gyroscope are retrieved via I2C. The measurements are converted into physical units and timestamped to ensure accurate time integration.

*2. Angle Computation*

The accelerometer data are used to compute a tilt angle based on the direction of gravity. The gyroscope data are integrated over time to obtain a second estimate of the tilt angle.

*3. Kalman Filter-Based Sensor Fusion*

A Kalman Filter is applied to combine the accelerometer and gyroscope estimates. The gyroscope data are used for prediction, while the accelerometer data are used for correction. This process reduces noise and compensates for drift.

*4. Output and Logging*

The filtered tilt angle is output through a serial interface and recorded in CSV format for further analysis. This data is later used for quantitative evaluation, including noise, drift, and accuracy analysis. The resulting data will be plotted using R for a more visual demonstration of the experiments

### 2.4 Implementation Details

The system is implemented using an embedded programming environment compatible with the RP2040 platform. Real-time constraints are considered to ensure that all computations, including sensor reading and Kalman filtering, are completed within each sampling interval.

A consistent sampling interval is maintained using a timing control mechanism, and the time step (dt) is calculated dynamically to improve integration accuracy. Basic sensor calibration is performed to reduce bias, particularly for the gyroscope, which is critical for minimizing drift.

### 2.5 Reproducibility

To facilitate reproducibility and further development, the full hardware configuration and source code are publicly available. Refer to Appendix.

## 3. Methodology

### 3.1 Accelerometer-Based Angle Estimation

The accelerometer measures the projection of gravitational acceleration along its sensing axes. When the sensor is stationary or moving at a constant velocity, the measured acceleration can be used to estimate the tilt angle relative to gravity.

For a single-axis tilt estimation, tilt angle can be computed using the ratio of the two acceleration components. Mathematically, it can be expressed as:

$$\theta_{acc} = arctan(\frac{a_y}{a_z}) \qquad [1]$$

Where $a_y$ and $a_z$ are the accelerometer readings along the corresponding axes.

Although this method provides an absolute reference based on gravity, it is highly sensitive to noise and external disturbances, resulting in significant short-term fluctuations in the estimated angle.

### 3.2 Gyroscope-Based Angle Estimation

The gyroscope measures angular velocity, which can be integrated over time to estimate the tilt angle. The angle at time step $t$ can be computed as:

$$\theta_{gyro}(t) = \theta_{gyro}(t - 1) + \omega \cdot dt \quad [2]$$

where ω is the angular velocity measured by the gyroscope, and $dt$ is the time interval between successive measurements.

This approach provides smooth and responsive angle estimation in the short term. However, due to sensor bias and noise, the integration process accumulates errors over time, leading to drift in the estimated angle.

### 3.3 Kalman Filter for Sensor Fusion

To address the limitations of individual sensors, a Kalman Filter is used to combine accelerometer and gyroscope measurements. The filter estimates the true tilt angle by recursively updating its state based on prediction and correction steps.

#### *3.3.1 Prediction Step*

In the prediction step, the gyroscope measurement is used to estimate the current angle based on the previous state:

$$\theta_k^- = \theta_{k-1}^- + \omega \cdot dt \quad [3]$$

This prediction leverages the short-term smoothness of the gyroscope data.

#### *3.3.2 Update Step*

In the update step, the accelerometer-based angle is used to correct the predicted value:

$$\theta_k^- = \theta_{k-1}^- + K(\theta_{acc} - \theta_k^-) \quad [4]$$

where $K$ is the Kalman gain, which determines the relative weighting between the prediction and the measurement.

The Kalman gain is dynamically adjusted based on the estimated uncertainty of the system, allowing the filter to rely more on the gyroscope during rapid motion and more on the accelerometer during stable conditions.

### 3.4 Filter Characteristics

The Kalman Filter effectively combines the complementary strengths of the two sensors:

- The gyroscope provides smooth short-term estimation but suffers from drift
- The accelerometer provides long-term reference but is noisy

By integrating both sources, the filter produces an angle estimate that is both stable and responsive.

In practical implementation, the filter parameters are tuned to balance noise reduction and responsiveness. Proper tuning is essential to achieve optimal performance under different motion conditions.

### 3.5 Implementation Considerations

To ensure accurate estimation in an embedded environment, several practical considerations are addressed:

- **Time step accuracy:** The time interval $dt$ is measured dynamically to improve integration precision
- **Sensor calibration:** Initial bias in the gyroscope is estimated and compensated
- **Numerical stability:** The filter is implemented using a simplified model suitable for real-time execution on a microcontroller
- **Computational efficiency:** The algorithm is optimized to run within the fixed sampling interval without introducing latency

## 4. Experimental Setup

The following 5 experiments will be used to test our Kalman filter methodology, by putting the filter in both static and kinetic experiments alongside the other two signals: accelerometer and gyroscope. Data will be collected over a short term period (10-30 seconds) or a long term span(120 seconds). These data will then be analyzed to determine whether the Kalman filter-infused algorithm is able to generate a smoother tracking by combining the strengths of the accelerometer and gyroscope signal.

The performance of the 3 signals will be evaluated based on the extent of fluctuations of their detected values (measured using the standard deviation of the data set) over each measured period. For the exact methodology and exceptions while calculating the standard deviation, refer to *5.1.2 Experiment 1 Result Analysis*.

## 4.1 Experiment 1 Setup

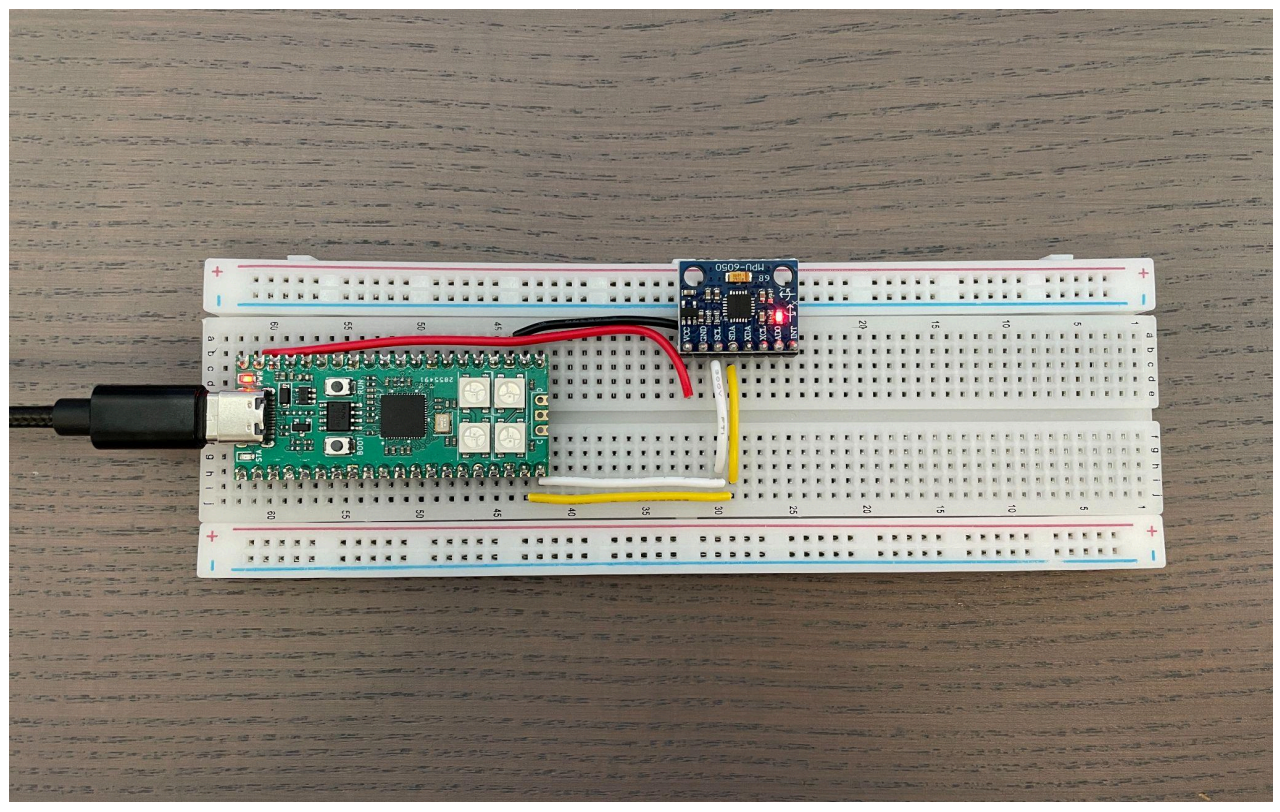

*Fig 3: The hardware setup of Experiment I on a breadboard.*

### *4.1.1 Setup description*

A Raspberry Pi RP2040 connected to a MPU6050 sensor on a standard breadboard. The RP2040 is then connected to Thonny in which the code will be run.

### *4.1.2 Experiment description*

The purpose of this experiment is to evaluate the short-term noise characteristics of the MPU6050 sensor under stationary conditions and to compare the stability of different angle estimation methods in the sensor.

The sensor is placed on a flat surface and kept completely still during the experiment. Data are continuously recorded over a fixed duration, including:

Accelerometer-based angle estimation ($angle_{acc}$)

Gyroscope-integrated angle estimation ($angle_{gyro}$)

Kalman filter fused angle ($angle_{kalman}$)

By analyzing the extent of variation of these three measurements, this experiment aims to identify the effectiveness of a Kalman Filter-infused algorithm as a means to reduce noise from the individual components of the accelerometer and the gyroscope.

## 4.2 Experiment 2 Setup

### *4.2.1 Setup description*

Experiment 2 will follow the same setup as Experiment 1. For setup diagrams and instructions refer to section *4.1.1* and Fig III.

### *4.2.2 Experiment description*

The purpose of this experiment is to evaluate the long-term characteristics of the MPU6050 sensor under stationary conditions and to investigate the limitations of regular angle estimation methodologies over extended periods.

The rest of the experiment is kept constant as experiment 1. Refer to section *4.1.2*

## 4.3 Experiment 3 Setup

To facilitate moving the sensor at an incline, the breadboard is attached onto a digital protractor angle finder. The protractor would be opened at different angles for the sensor to collect data upon. The protractor has extreme precision and lays a strong foundation for collecting data at an angle.

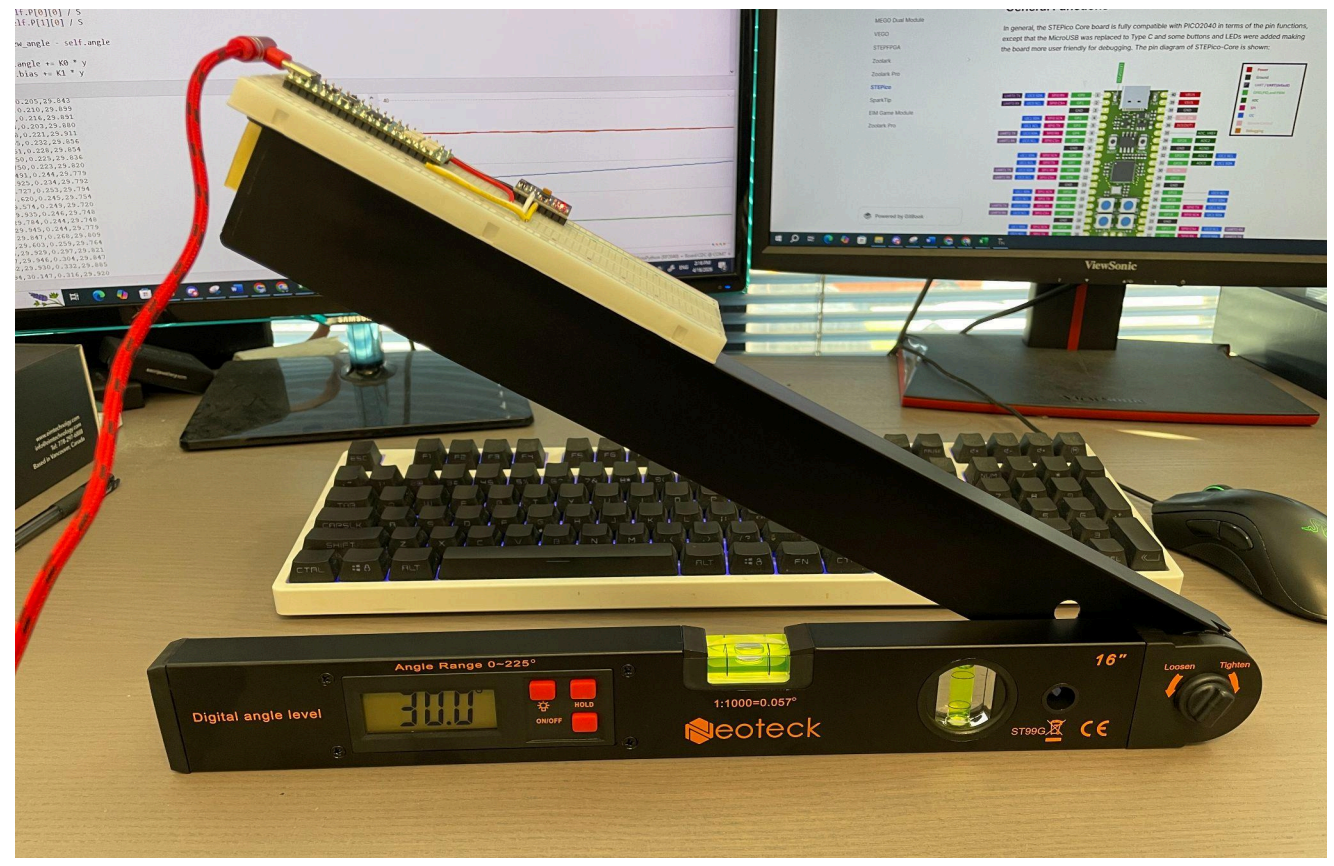

*Fig 4: The setup of Experiment 3 with protractor-determined angle at 30.0°*

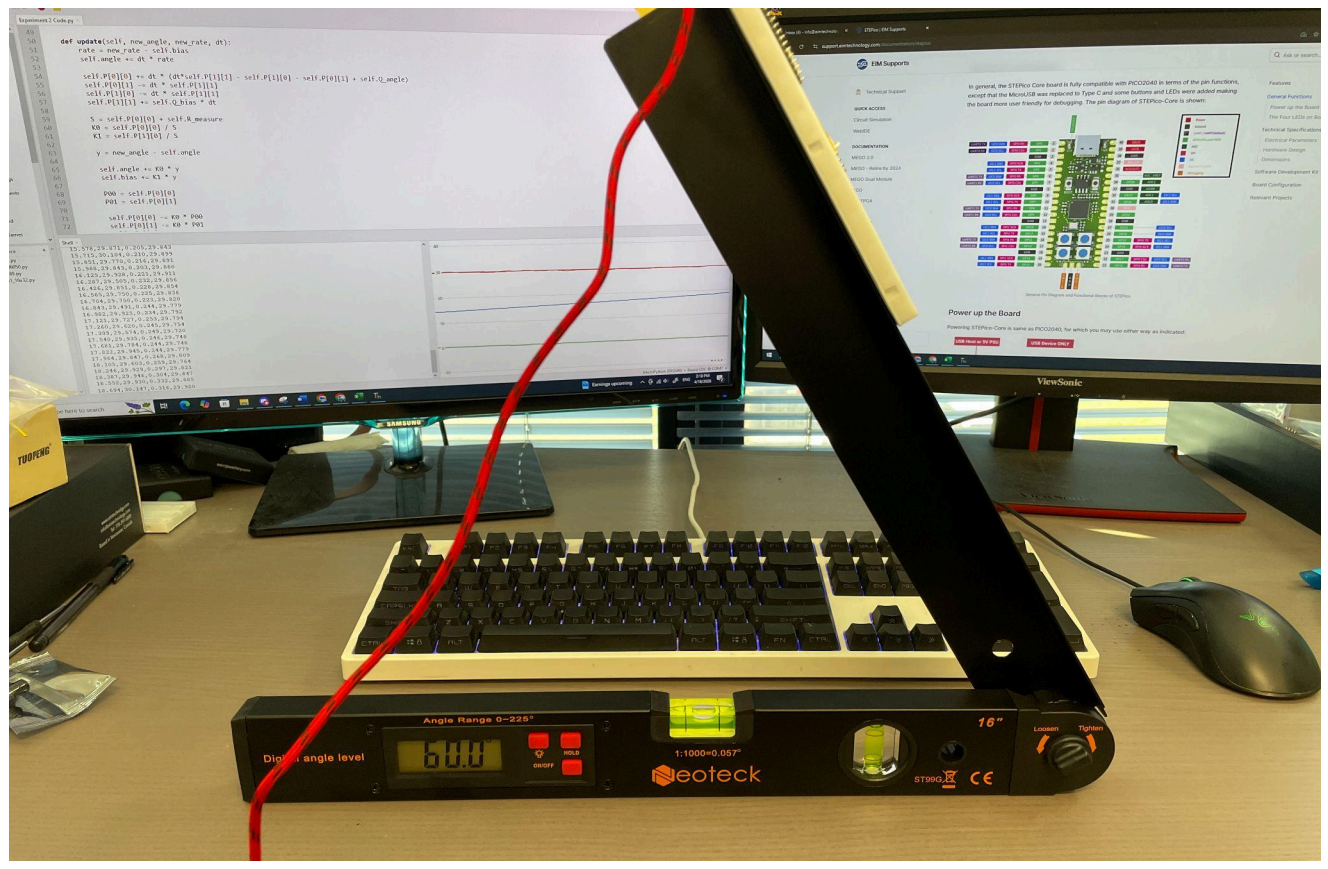

*Fig 5: The setup of Experiment 3 with protractor-determined angle at 60.0°*

### *4.3.1 Setup description*

Experiment 3 will follow the same hardware setup as Experiments 1 and 2. For setup information refer to section *4.1.1.* This experiment will also feature a digital protractor angle finder, which the breadboard is tied to.

### *4.3.2 Experiment description*

The purpose of this experiment is to evaluate the accuracy of the estimation system under known,

fixed-angle conditions and to compare the performance of different estimation methods.

The MPU6050 sensor is placed at several predefined angles using external references using a leveling tool. At each angle, the sensor is kept completely stationary, and data is recorded over a short duration. The same signals are measured:

Accelerometer-based angle estimation ($angle_{acc}$)

Gyroscope-integrated angle estimation ($angle_{gyro}$)

Kalman filter fused angle ($angle_{kalman}$)

For this experiment, we will use $angle_{acc}$ and $angle_{kalman}$ as viable data, as a gyroscope finds difficulty in measuring angles at an incline. At each angle measured, an average is taken to determine the measured data vs. the assumed ideal value. This seeks to compare the accuracy of the accelerometer data as opposed to the data given by the Kalman algorithm.

### 4.4 Experiment 4 Setup

This experiment requires the sudden movement of the sensor between different angle measures in order to measure the speed of adjustment and its stability. Therefore, 3D printed angles are made as a guide for where to move the digital protractor angle finder, leading to the setup as shown below.

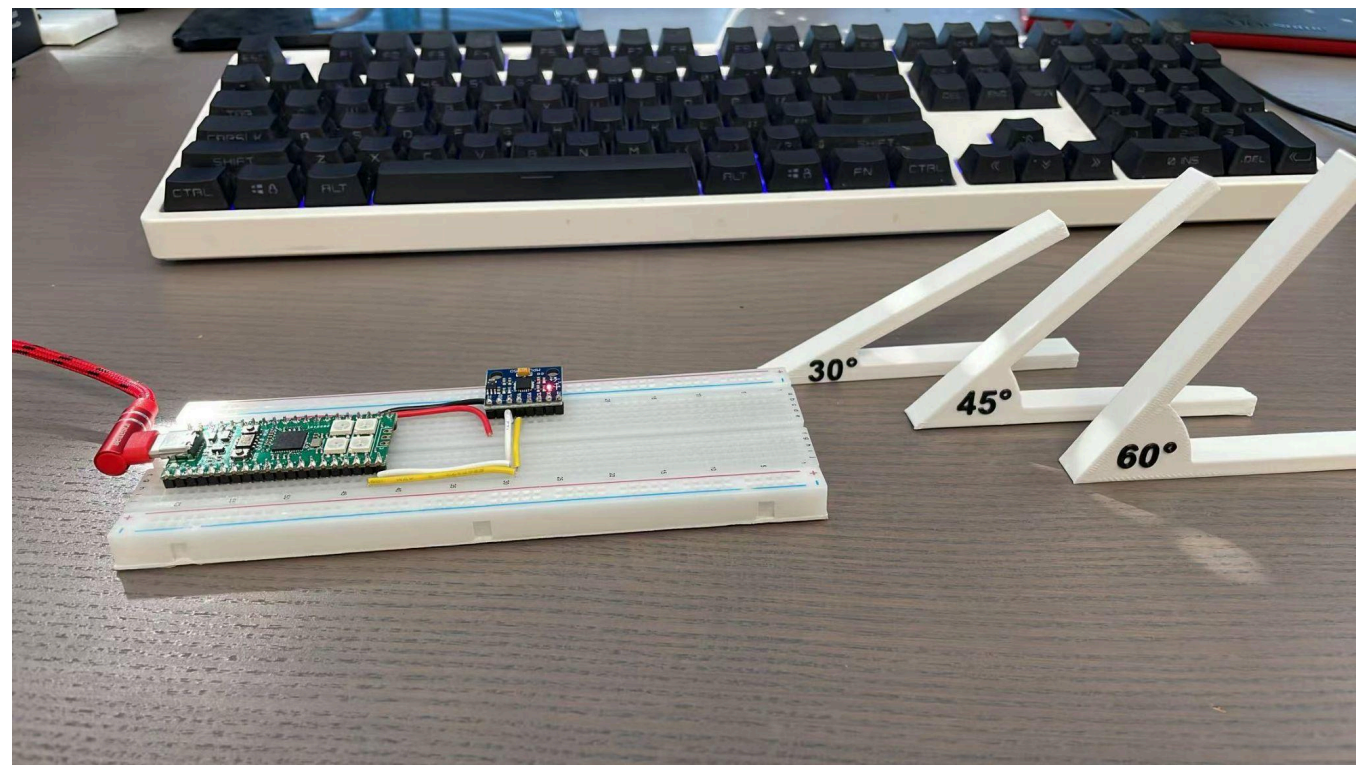

*Fig. 6: Hardware setup and external tools for Experiment 4.*

#### *4.4.1 Setup description*

The same hardware setup is kept. For setup information refer to section *4.1.1.* This experiment will also feature 3d printed angle measures that will be used as a reference to adjust the angle in the experiment.

#### *4.4.2 Experiment description*

This experiment introduces a set of fixed angle supports. These supports allowed the sensor to be quickly repositioned between predefined angles (0°, 30°, 45°, and 60°), enabling near instantaneous transitions followed by stable holding periods.

This setup reduces uncertainties caused by slow manual adjustment and ensures more well defined steps like changes in orientation. Consequently, it allows for clearer observation of system response, including responsiveness, stability, and noise characteristics of the accelerometer, gyroscope, and Kalman filter outputs.

The recorded signals during this experiment includes:

Accelerometer based angle estimation ($angle_{acc}$)

Gyroscope integrated angle estimation ($angle_{gyro}$)

Kalman filter fused angle ($angle_{kalman}$)

The purpose of this experiment is to test the algorithm's ability to adjust under rapid and sudden movements, while attempting to filter out introduced noises due to inconsistencies in the moving process.

### 4.5 Experiment 5 Setup

This experiment will differ from the setups of Experiments 1-4 as it requires data collection in a rolling motion, therefore the microcontroller cannot be connected to Thonny via a wire. For this experiment, we will use a MEGO portable battery as our power source, paired with a green LED to determine when data collection has begun. The RP2040 will automatically run the program once the battery is opened.

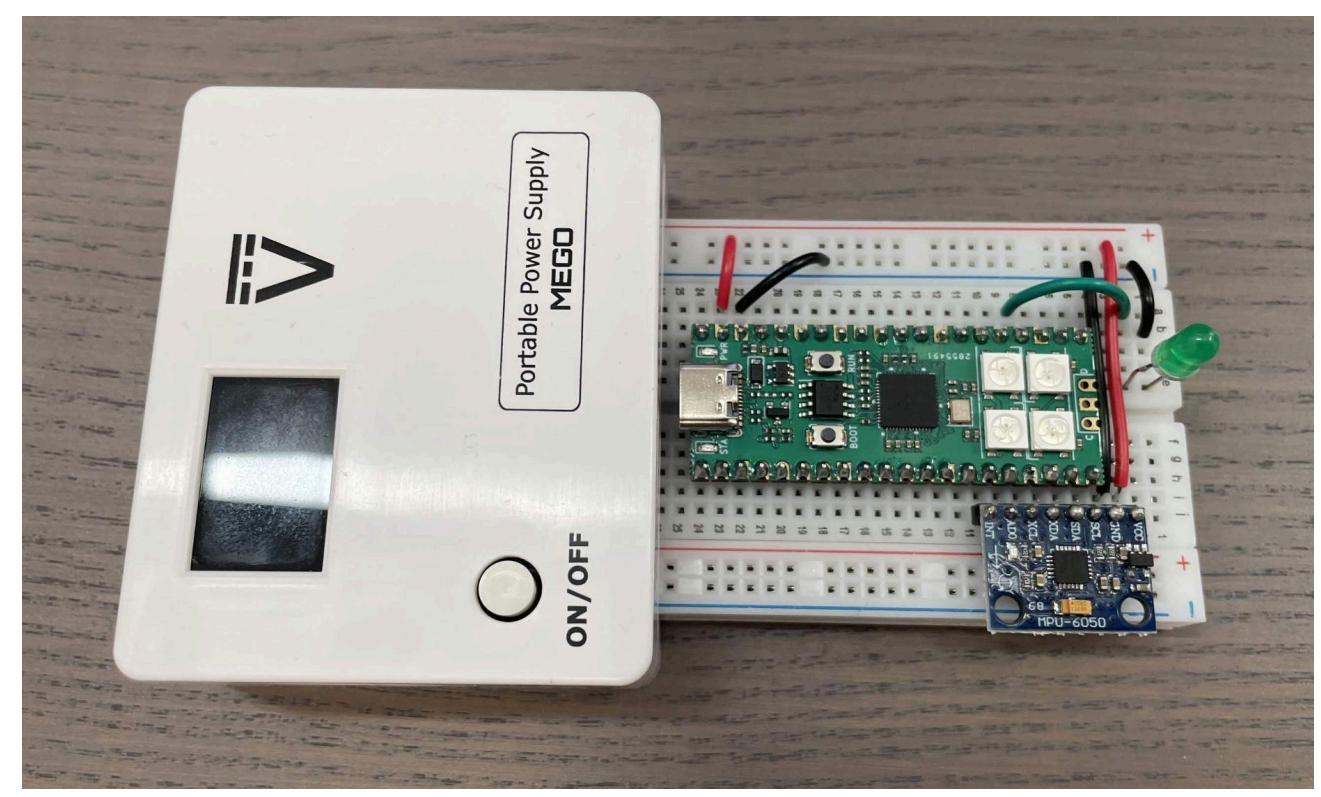

*Fig. 7: New portable supply and hardware setup for Experiment 5.*

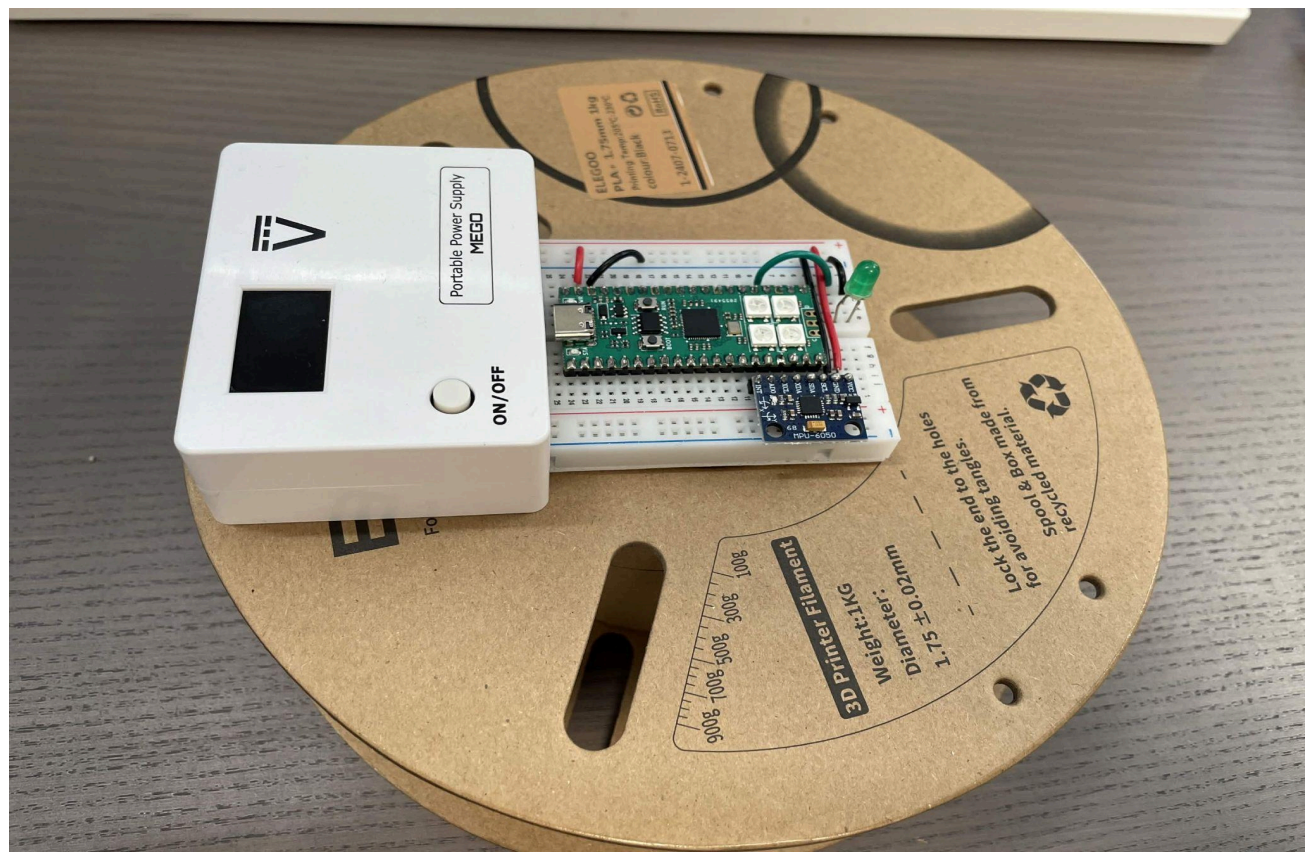

*Fig. 8: The circuit system on the wheel which will be rotated to gather rotational data.*

#### *4.5.1 Setup description*

A Raspberry Pi RP2040 connected to a MPU6050 sensor on a shortened breadboard, using a MEGO Portable Battery Supply. A Green LED connected to the to indicate when data has started to collect. The entire contraption is then glued to a 3d printer filament holder in order to conduct smooth circular motion.

#### *4.5.2 Experiment description*

This experiment seeks to test the stability of the Kalman Filter-infused algorithm under circular periodic motion. The same three signals: $angle_{acc}$, $angle_{gyro}$, and $angle_{kalman}$ are measured, and we will examine whether the signals provide a smooth, repeatable, and periodic tracking.

Due to the limited space available on the half breadboard, the arrangement of the portable power supply, microcontroller module, and MPU6050 sensor had to be optimized carefully. As a result, the MPU6050 communication pins were fixed to the following GPIO configuration:

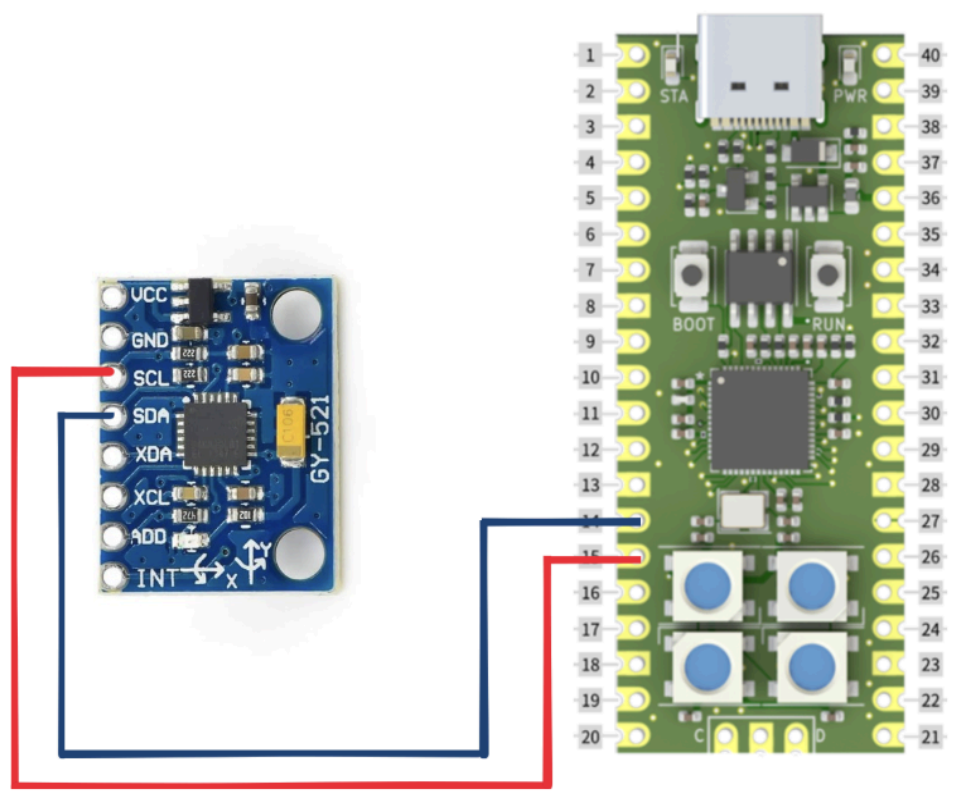

*Fig. 9: The GPIO configuration of the MPU 6050 to the RP2040 STEPico in Experiment 5*

A green LED indicator was introduced to provide visual feedback during data acquisition. The LED turns on only while the system is actively recording and saving data. When the LED is off, no data recording takes place. This feature helps ensure clear synchronization between experimental motion and the data collection period during rotational testing.

## 5. Quantitative Results and Analysis

### 5.1 Experiment 1

#### *5.1.1 Experiment 1 Graphical Results*

The following graphs (Figures X - XII) illustrate the short-term fluctuations of the 3 signals $angle_{acc}$, $angle_{gyro}$, and $angle_{kalman}$ across 3 repeated trials in the same environment. The experiments were conducted back to back to minimize any bias caused by non-human factors.

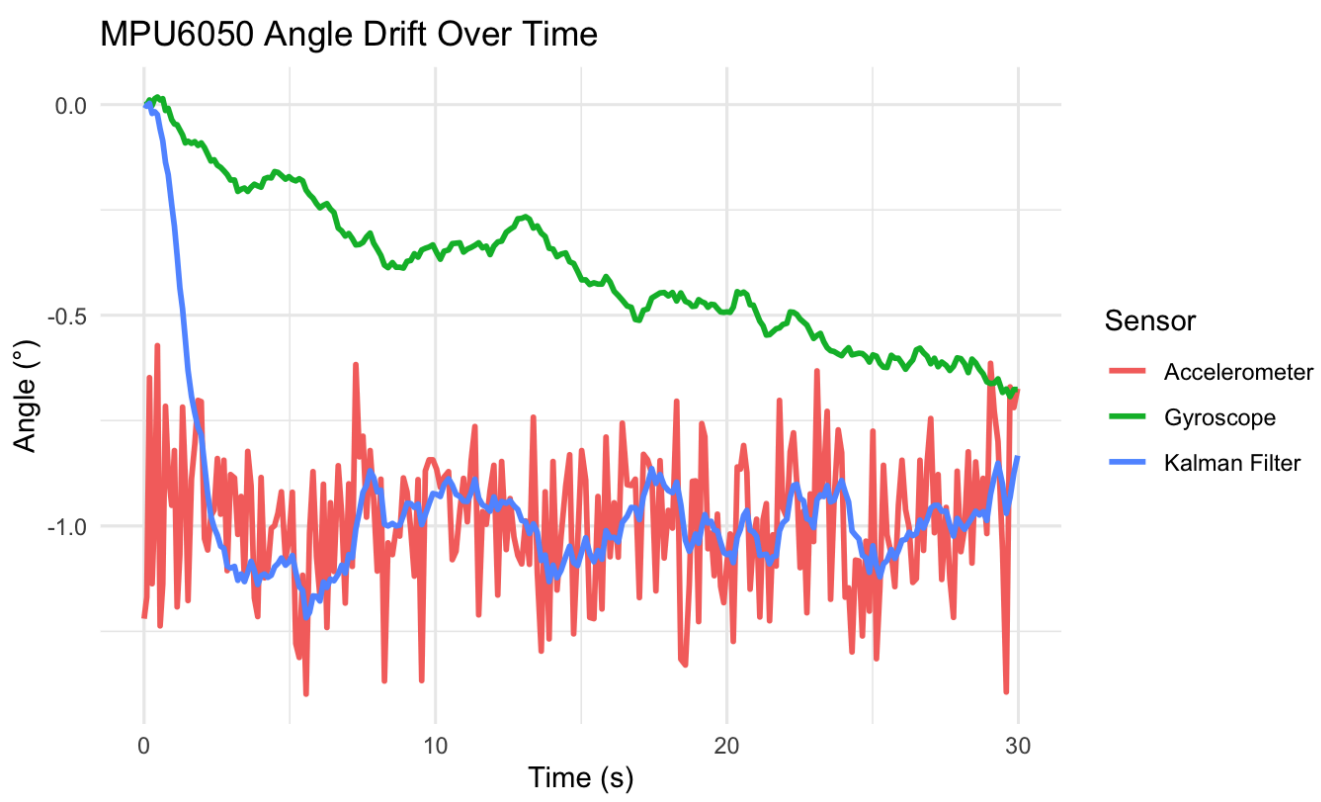

*Figure 10: Experiment 1 Trial 1 data*

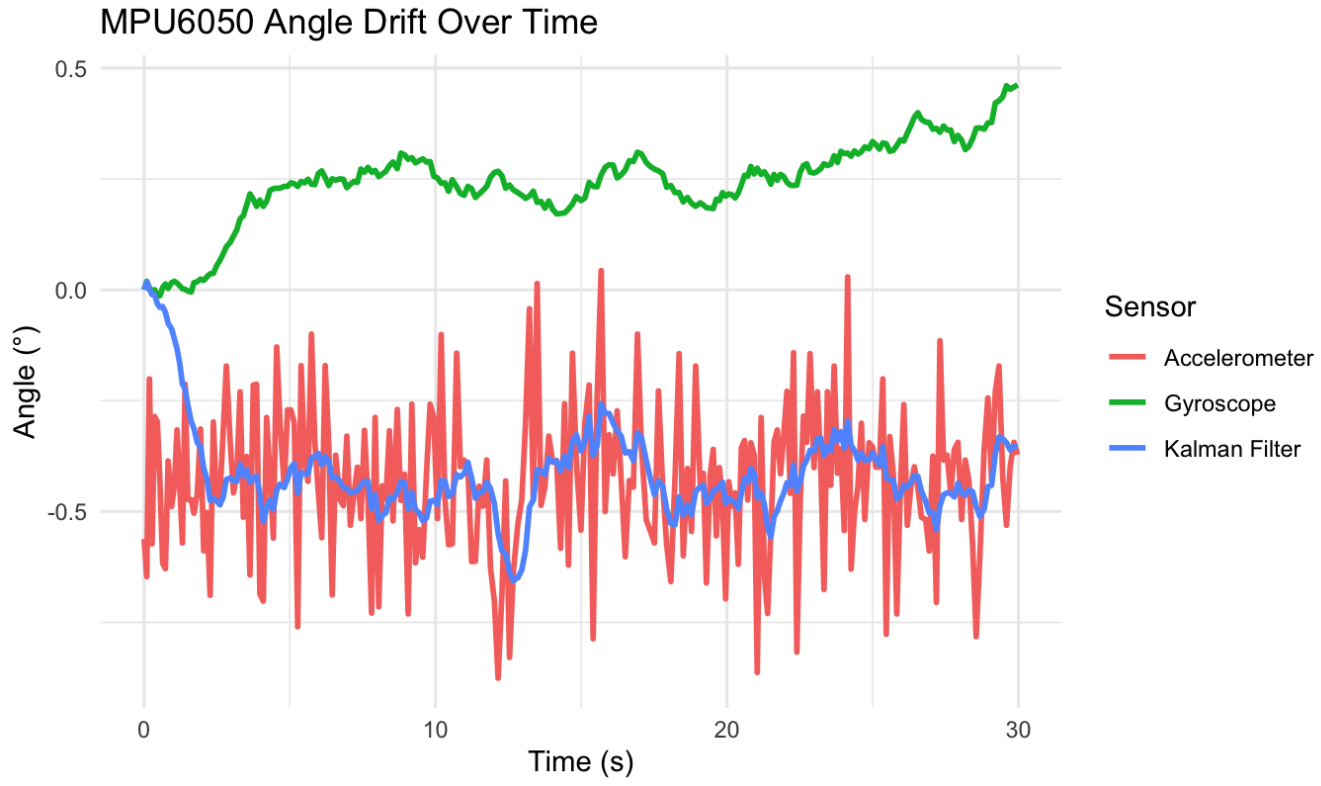

*Figure 11: Experiment 1 Trial 2 data*

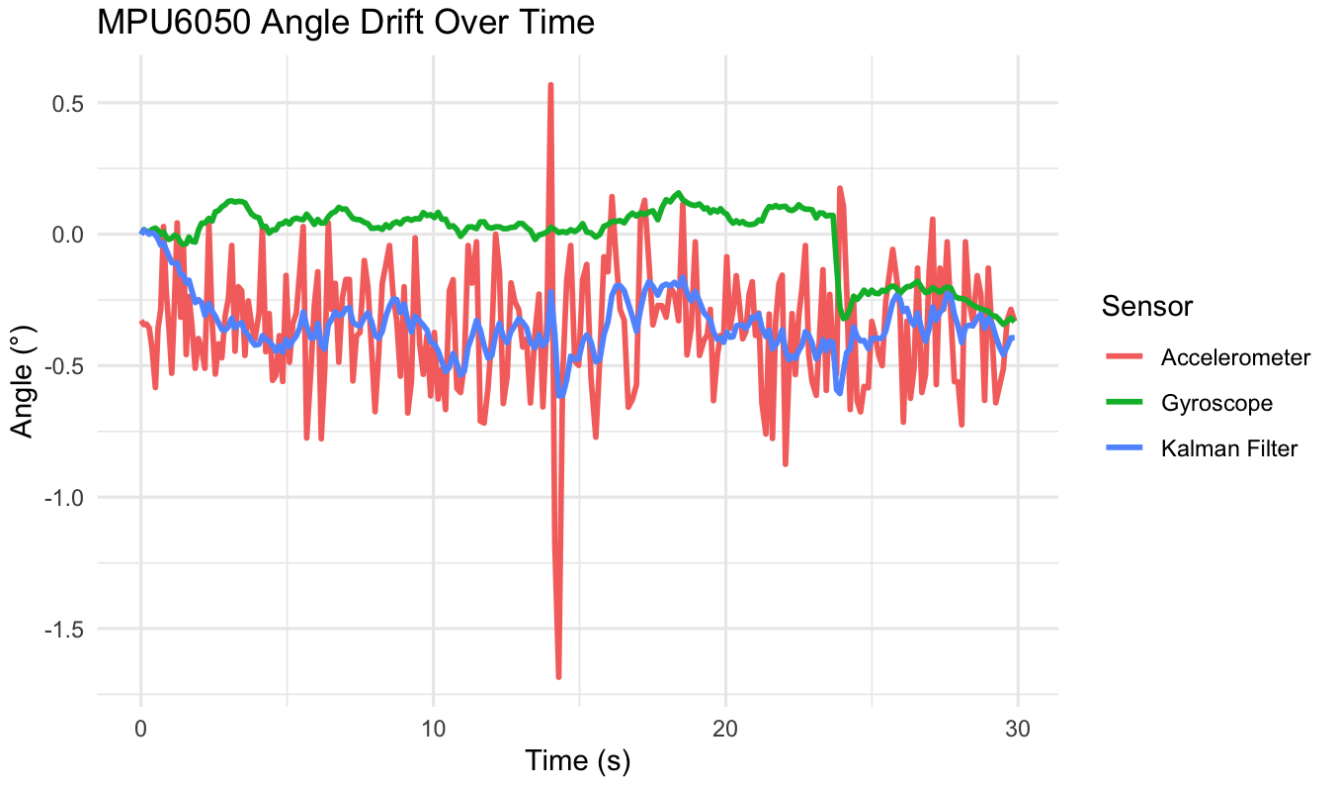

*Figure 12: Experiment 1 Trial 3 data*

### *5.1.2 Experiment 1 Result Analysis*

The experimental results show that the accelerometer-based angle exhibits noticeable fluctuations, with an average standard deviation of 0.192° across 3 trials, indicating significant measurement noise. The gyroscope-based angle is relatively smooth but still shows variability, with an average standard deviation of 0.136°.

After the Kalman filter adjusted to the true value*, the remaining data reached an average standard deviation of 0.075°, demonstrating highly improved stability. This confirms that the Kalman filter effectively reduces noise while maintaining a consistent angle estimate.

For trial 3 specifically, it is well noted that the accelerometer suffered a significant spike at around 14 seconds, which is most likely due to a non-human factor. It can be noted that the Kalman filter algorithm fluctuated very little under that influence.

** Because of a noticeable adjustment period of the Kalman algorithm that generates significant outliers, the standard deviation is calculated using the data points that are not considered an outlier by the standard statistics methodology, where an outlier is considered as a point that is less than Q1-1.5(IQR) or more than Q3+1.5(IQR).*

## 5.2 Experiment 2

### *5.2.1 Experiment 1 Graphical Results*

The following graphs (Figures XIII - XV) illustrate the long-term fluctuations of the 3 signals $angle_{acc}$, $angle_{gyro}$, and $angle_{kalman}$ across 3 repeated trials in the same environment. The experiments were conducted back to back to minimize any bias caused by factors concerning changes in the environment.

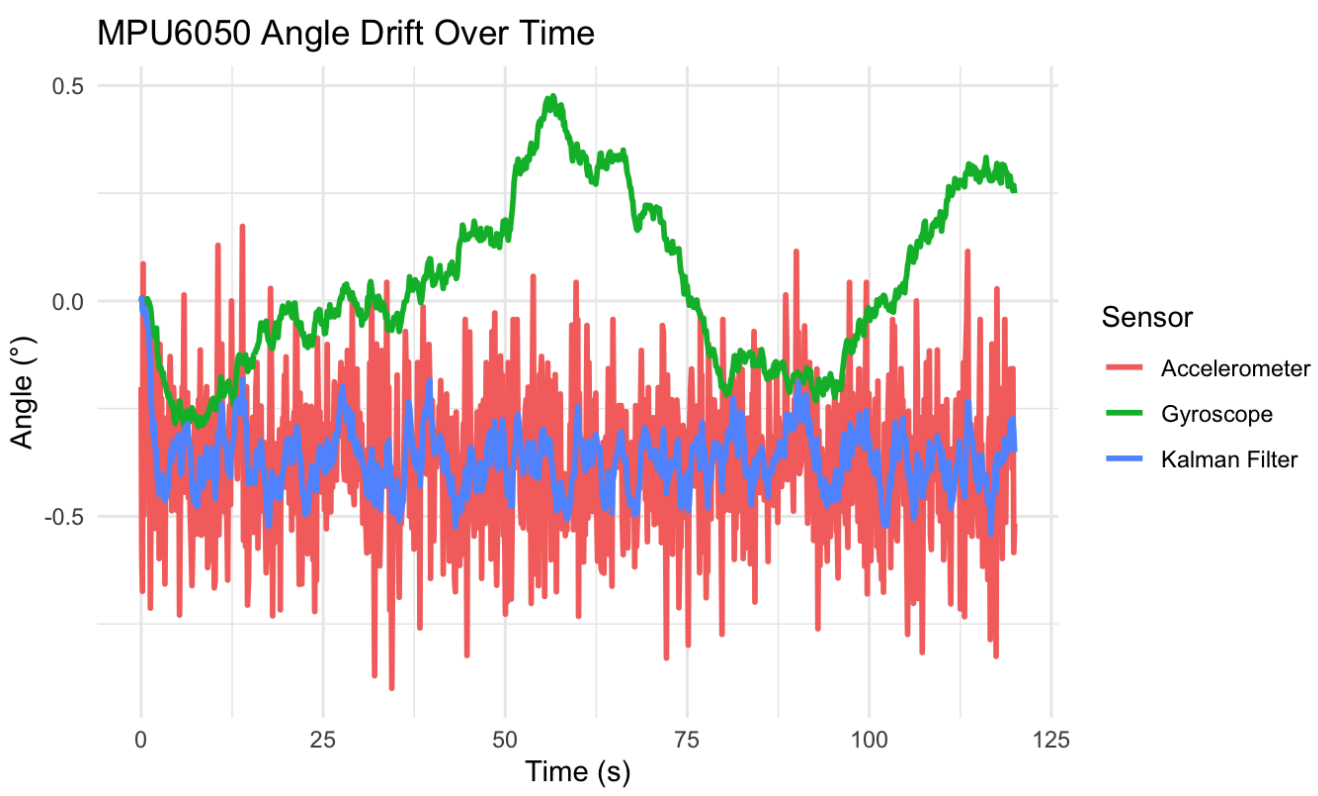

*Figure 13: Experiment 2 Trial 1 data*

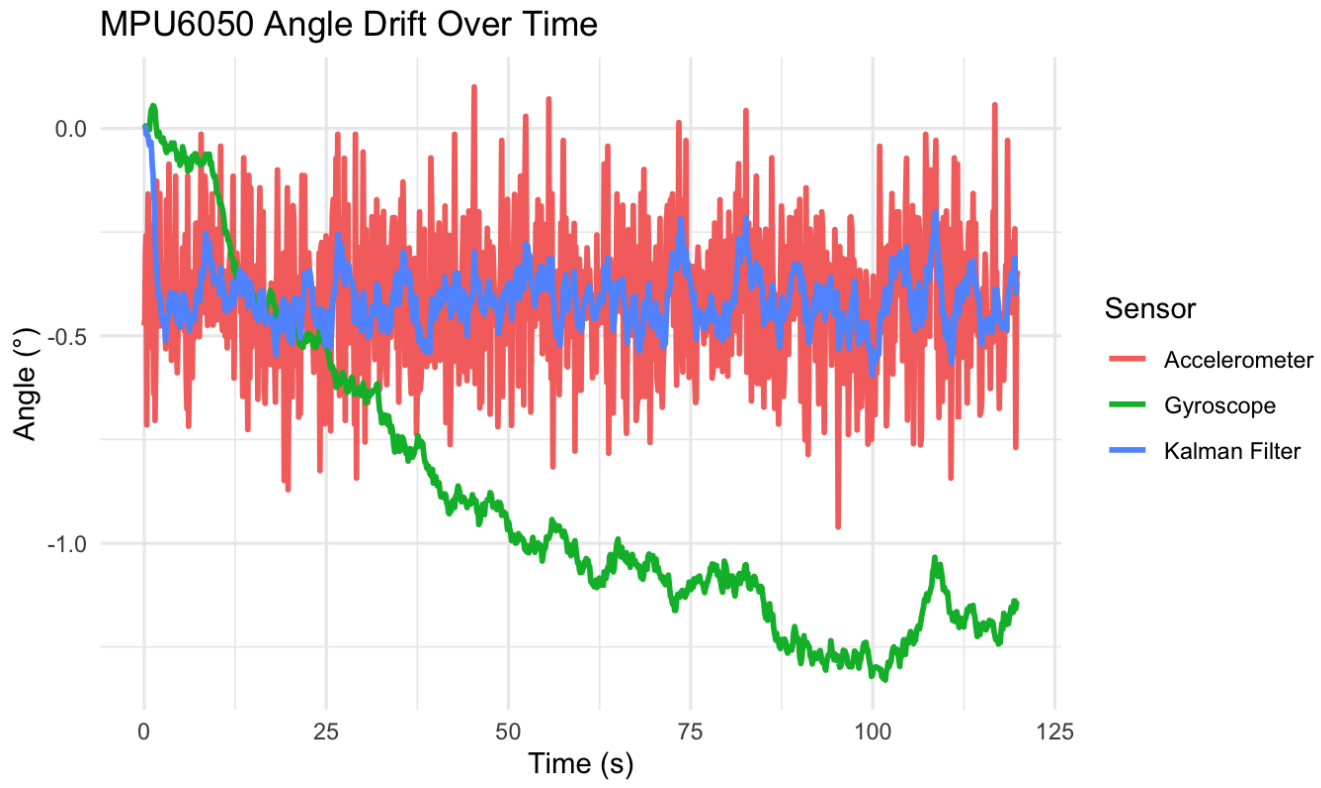

*Figure 14: Experiment 2 Trial 2 data*

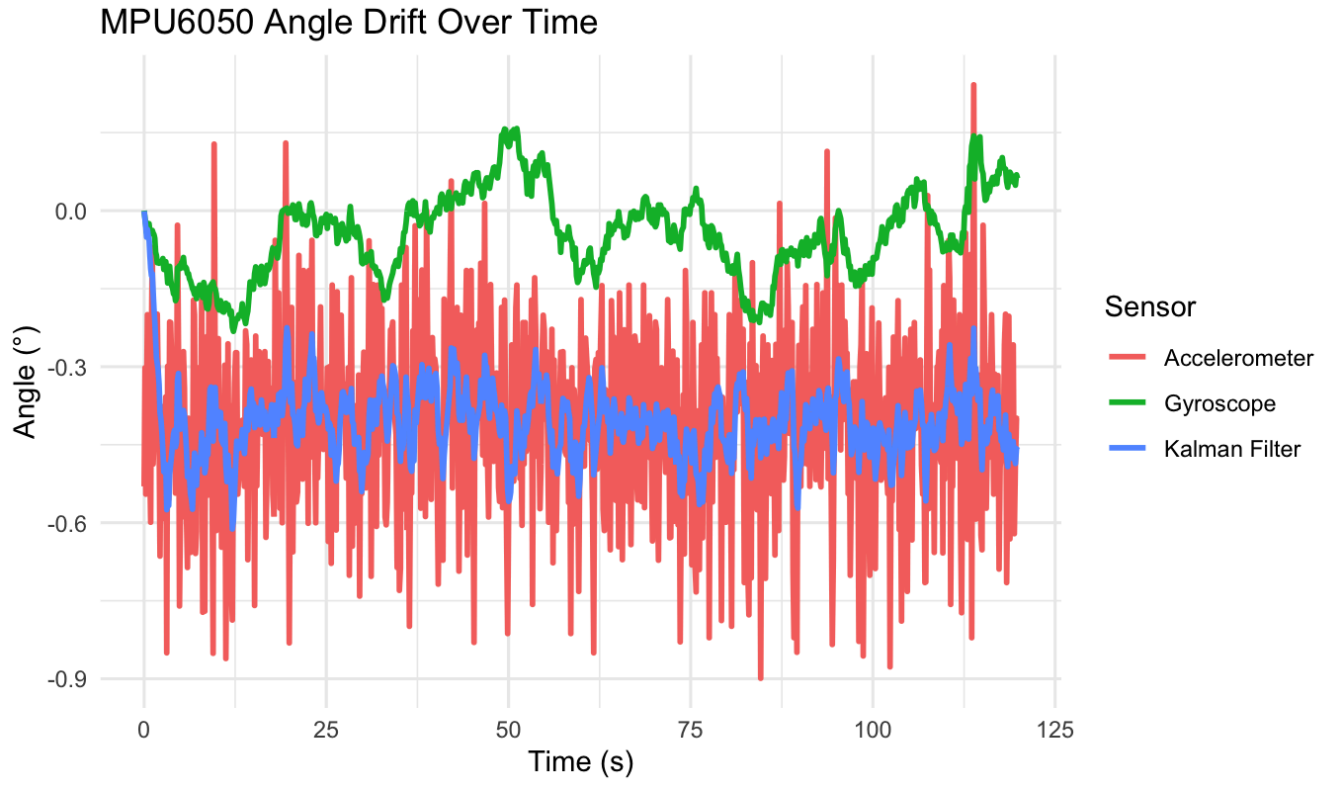

*Figure 15: Experiment 2 Trial 3 data*

### *5.2.2 Experiment 2 Result Analysis*

This 2 minute long-term tracking of the three signals has shown a trend similar to trial 1: While the accelerometer and gyroscope had an average standard deviation of 0.172° and 0.217°, the Kalman algorithm again showed its long term tracking prowess by recording an average standard deviation of 0.063° after initial adjustment.

Another point worth noting is that the long term drift of the gyroscope became quite lucid in this experiment; it measurement drifts in undesired directions and received a standard deviation higher than angle $angle_{acc}$ for the first time. This proves that the gyroscope is generally troublesome during long term tracking as compared to the accelerometer.

## 5.3 Experiment 3

### *5.3 Experiment 3 Graphical Results*

The following graphs (Figures XVI-XVII) are used to demonstrate the extent of deviations from the actual value of the accelerometer and Kalman filter-infused algorithm in set, non-zero angle conditions. Note that the gyroscope data will not be used for this experiment as a gyroscope cannot measure data if the sensor is at an incline.

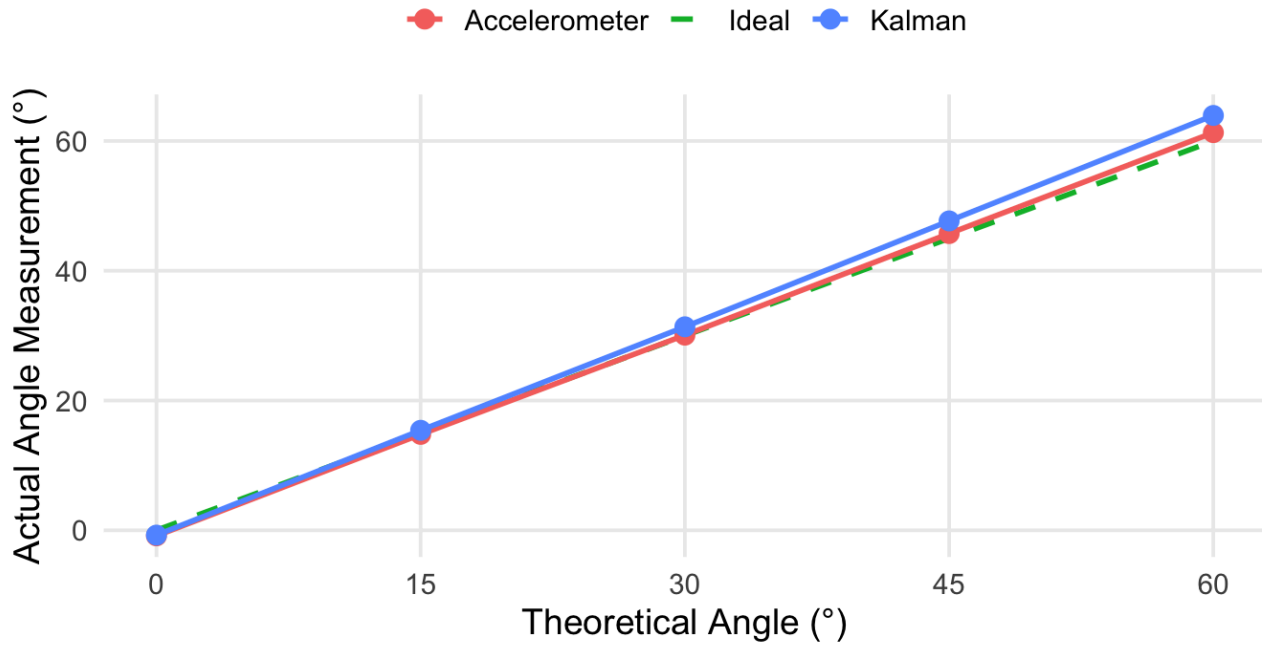

*Figure 16: Experiment 3 Trial 1 data*

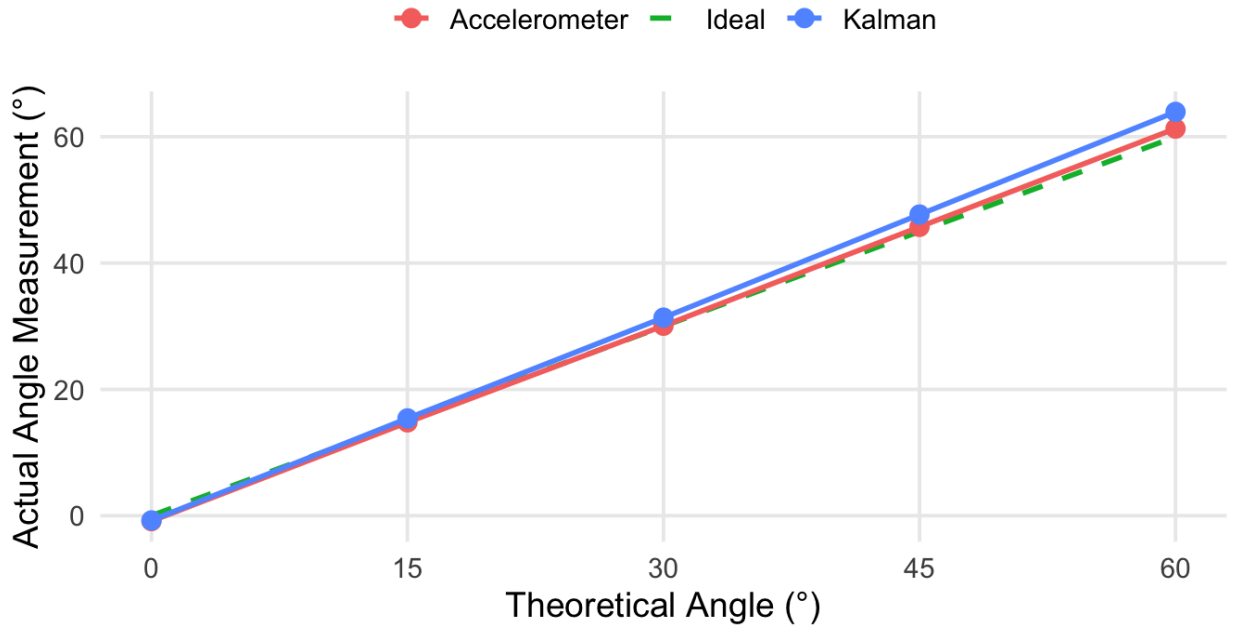

*Figure 17: Experiment 3 Trial 2 data*

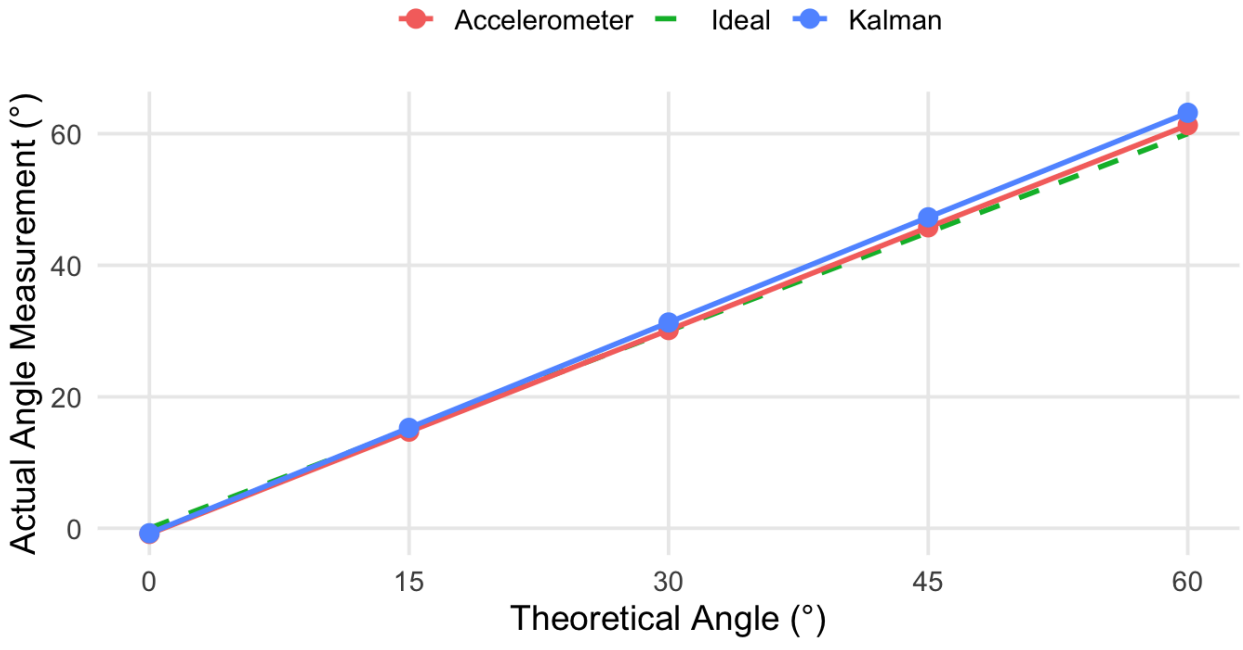

*Figure 18: Experiment 3 Trial 3 data*

### *5.3.2 Experiment 3 Result Analysis*

Experiment 3 presents data that are very consistent across the three trials, most likely due to the methodology of taking the average across 10 seconds for multiple angles. Again, due to the Kalman Filter's initial calibration, the outliers in the data set are again discarded.

However, in this experiment, the Kalman data drifts further away from the actual data as the angle increases. This can be explained by a process of overcalibration at each individual angle, which is not accounted for while outliers were discarded. For the graphs at each angle, refer to Appendix.

## 5.4 Experiment 4

### *5.4 Experiment 4 Graphical Results*

The following graphs (Figures XIX-XXI) will attempt to determine the stability of the three signals under abrupt movements of angles to 30°, 45°, and 60°. 3 trials will be conducted and the same pre-experiment conditions and environment are kept and assumed.

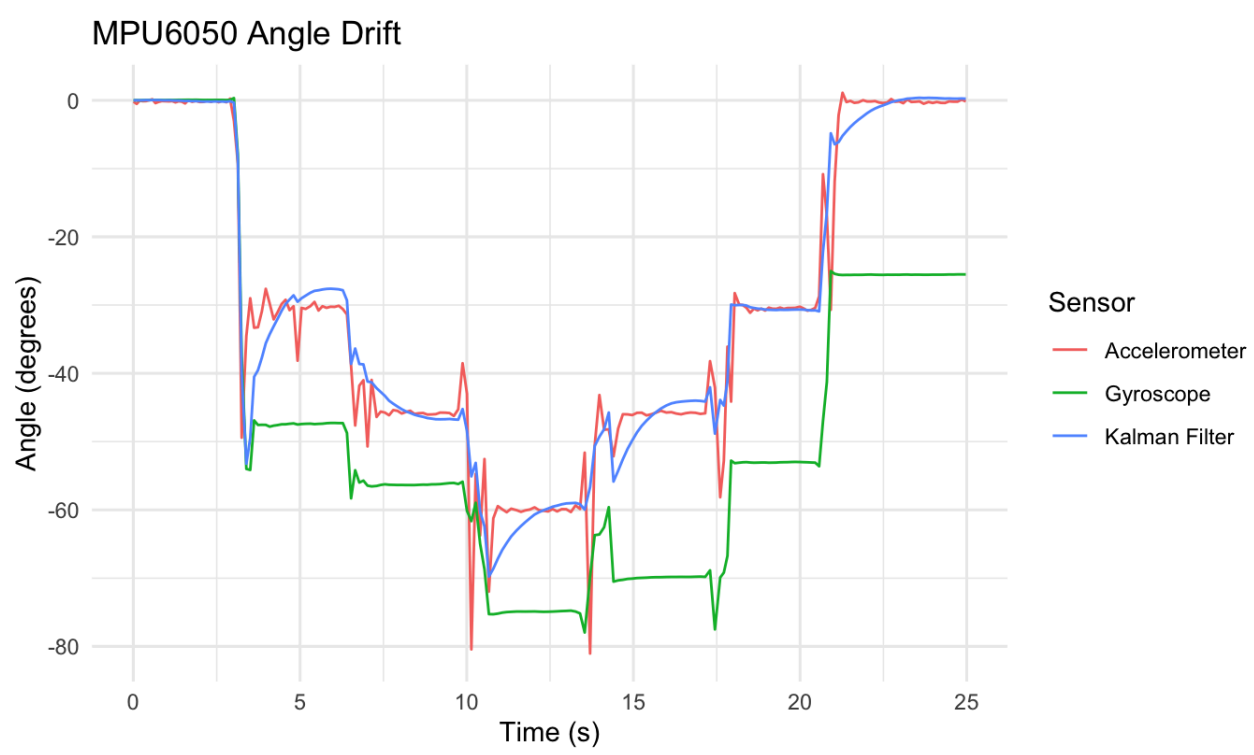

*Figure 19: Experiment 4 Trial 1 data*

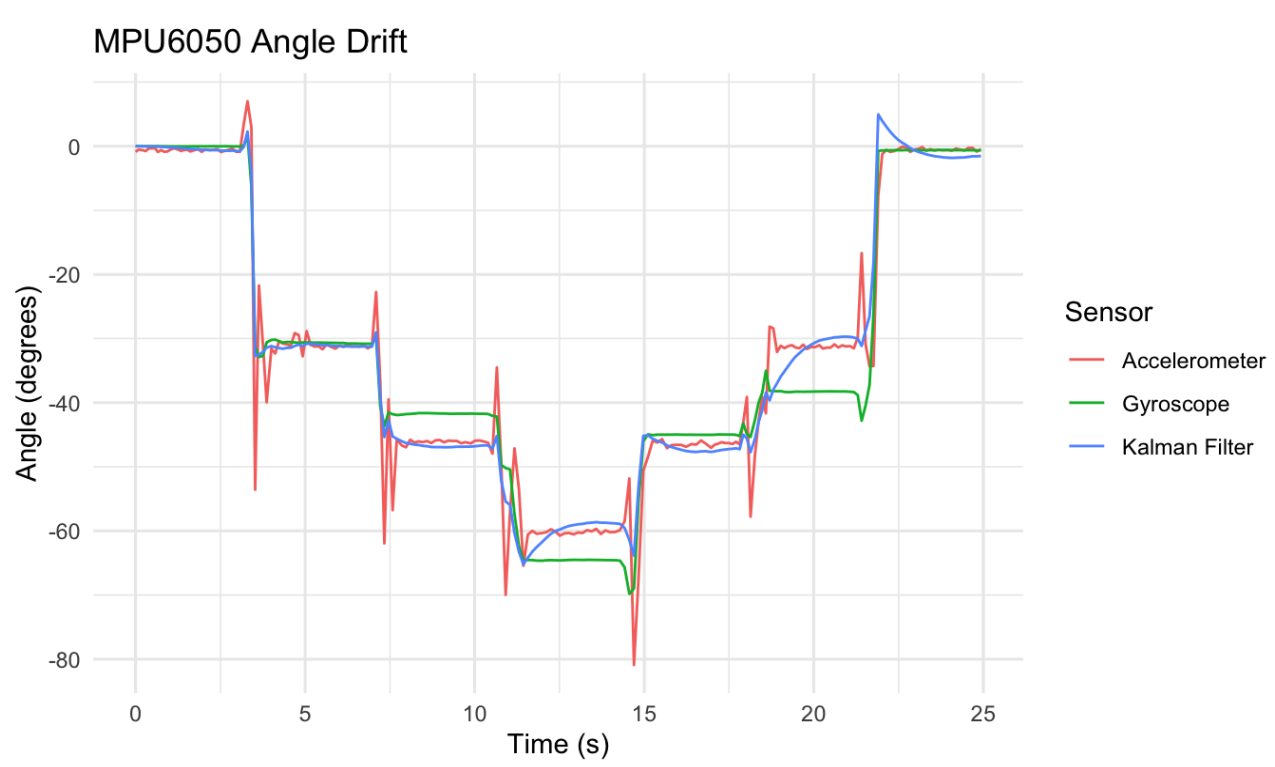

*Figure 20: Experiment 4 Trial 2 data*

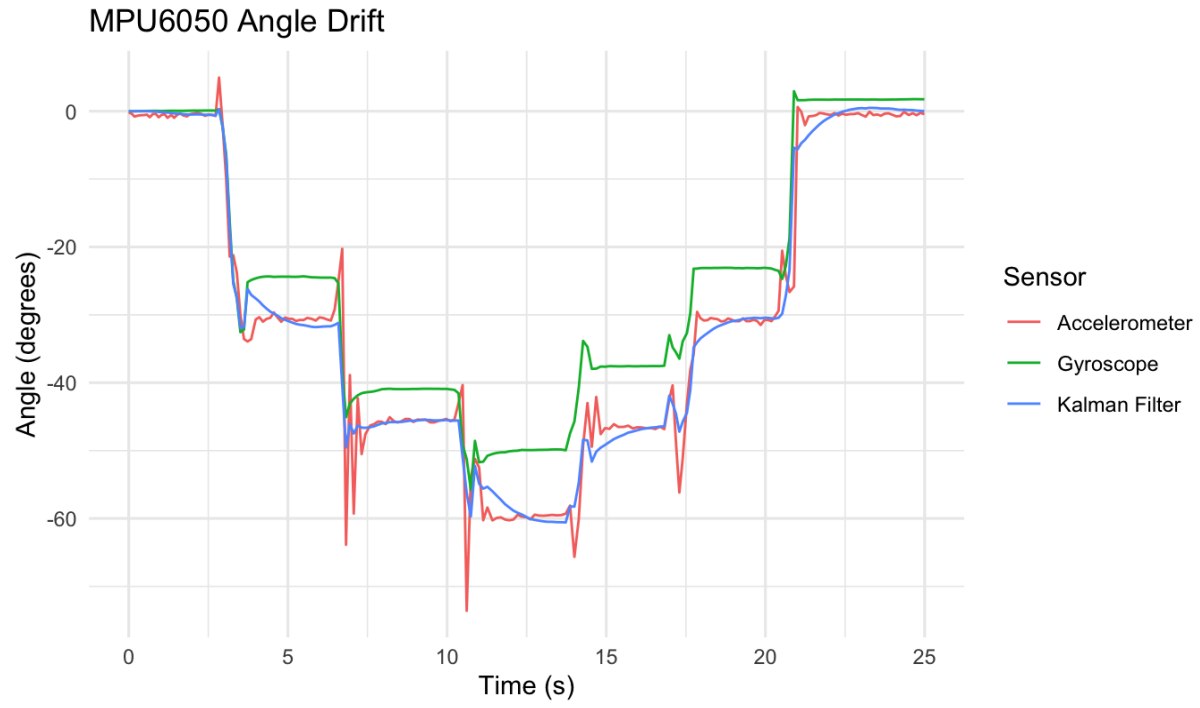

*Figure 21: Experiment 4 Trial 3 data*

### *6.4.2 Experiment 4 Result Analysis*

In this experiment, The Kalman algorithm demonstrated its ability to adjust to the actual value through rounding off noises. While the accelerometer and gyroscope data contains significant spikes that fluctuate between the actual value and those around it, the Kalman algorithm again presents a smooth tracking that is more accurate and less prone to errors in measurement.

Despite the smooth tracking of the Kalman filter, it is observed that at every sudden jolt the Kalman's line needs time to recover to the actual data. This delay may cause minor concerns, however once the calibration is finished, the signal steadies and presents a smooth line.

## 5.5 Experiment 5

### *5.5.1 Experiment 5 Graphical Results*

The following graphs (Figures XIII - XV) demonstrate the changes in the angle measured by the three sensors during periodic motion. 3 trials will be conducted and the same pre-experiment conditions and environment are kept and assumed. Again, examining a different gyroscope axis, particularly the axis perpendicular to the sensor board, produced little or no change due to inherent limitations with the gyroscope measurement.

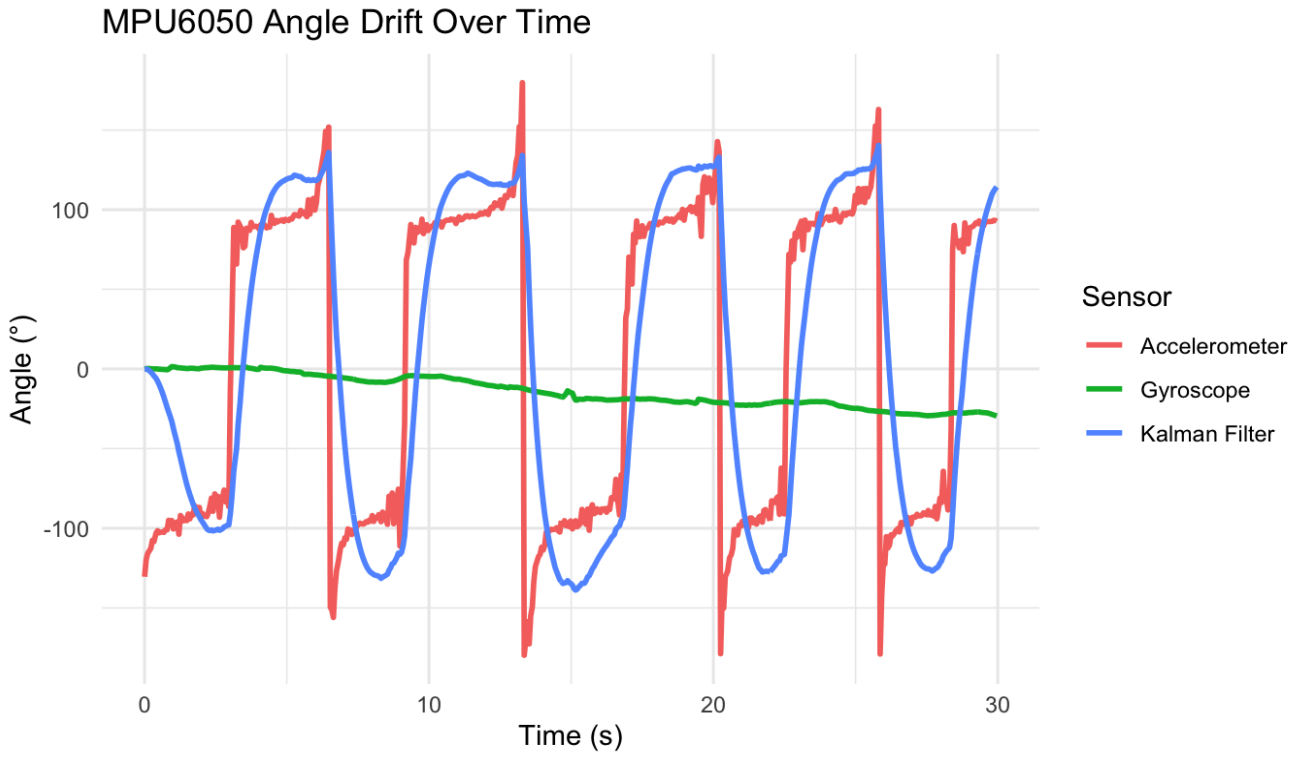

*Figure 22: Experiment 5 Trial 1 data*

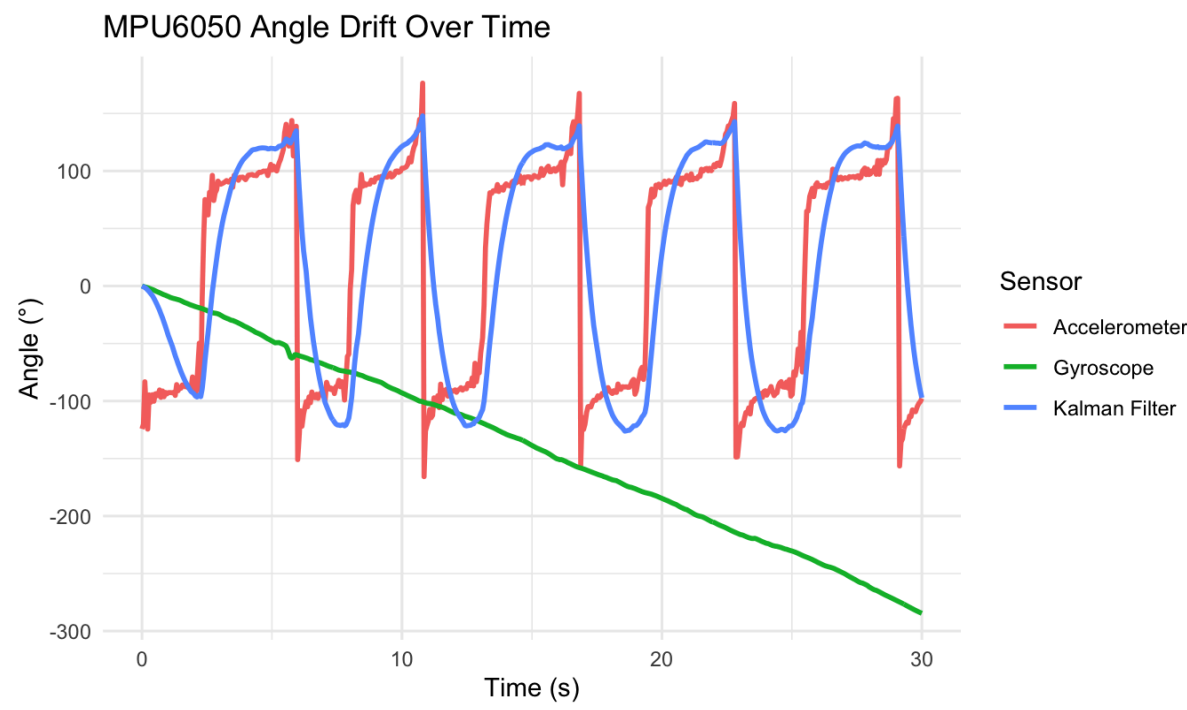

*Figure 23: Experiment 5 Trial 2 data*

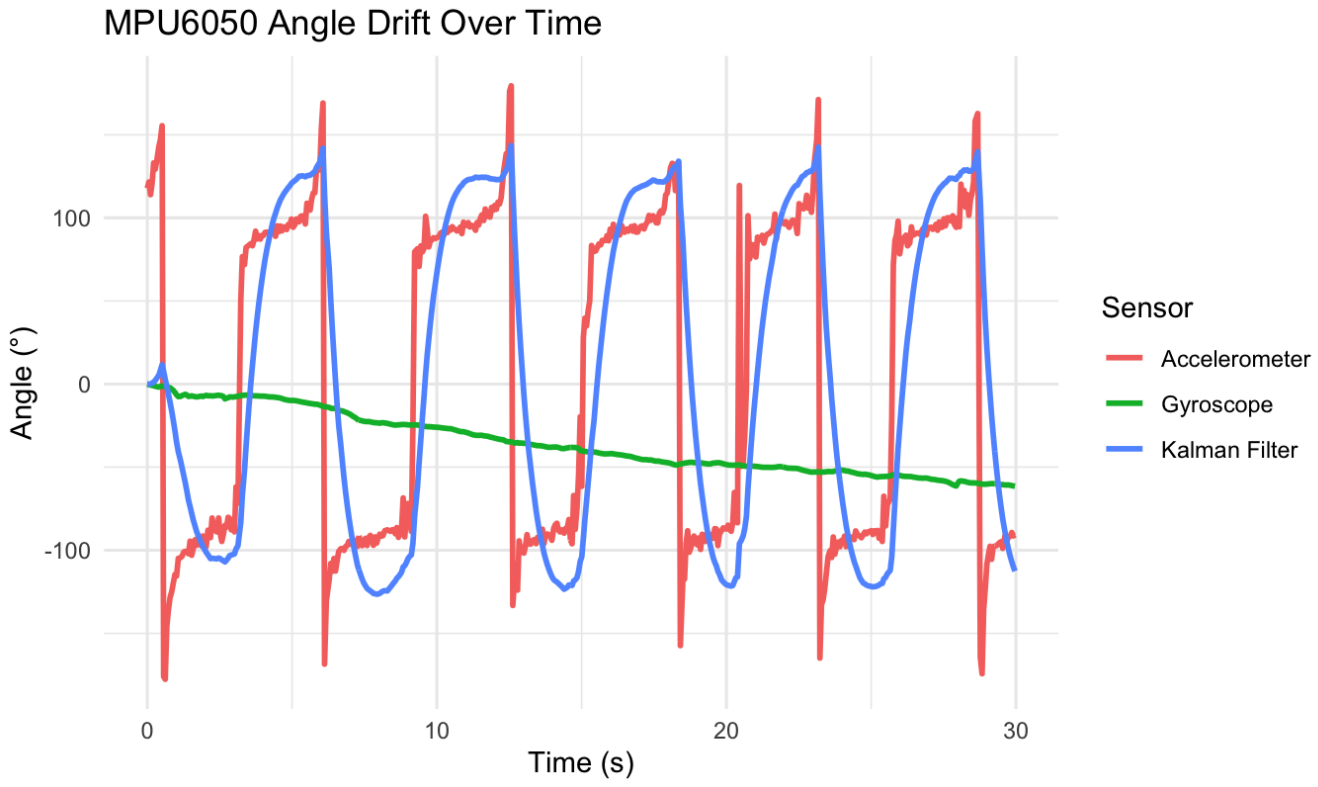

*Figure 24: Experiment 5 Trial 3 data*

### *5.5.2 Experiment 5 Result Analysis*

In each experiment, the gyroscope was not able to recognize circular motion, and therefore presented data that was flawed and did not reflect the true value.

The accelerometer's data resembles a sine wave, however there are many instances of spikes that may be caused by measurement noise. However the Kalman filter was able to smooth out the graph into almost a perfect sinusoidal curve. This again demonstrated that the Kalman algorithm is able to maintain smooth tracking even during rotations.

However, the source of noise is currently unknown; the noise could also be caused by the human's unintentional movement of the hand or an imperfect rolling process. This may cause problems if intentional microadjustments are made, but was omitted by the Kalman system as measurement noise.

## 6. Discussion

Through the 5 experiments, the Kalman filter-infused algorithm for object tracking produced a smooth result that allowed the noise during the tracking process to be massively reduced. The purpose of combining the strengths of the accelerometer and gyroscope data has been successfully achieved, generating results that are significantly better than the two original signals.

A potential improvement to this study is that the angle estimation is limited to a single axis, assuming motion within a planar configuration. The tilt angle is computed using accelerometer measurements based on a two-dimensional projection, and the Kalman filter is applied to fuse accelerometer and gyroscope data along this axis.

This approach simplifies the problem and allows clear observation of noise reduction and drift suppression effects. However, it does not account for full three-dimensional orientation, where cross-axis coupling and more complex motion dynamics may affect the estimation accuracy.

Future work may include extending this approach to full 3D orientation estimation using multi-axis sensor fusion methods, such as quaternion-based filters or complementary filters. This allows us to discover a new dimension in which the Kalman Filter-infused algorithm's utility is testified.

The MPU6050 must be attached to the breadboard, and while our experiments look to adjust the breadboard's angle, we cannot be sure that the sensor is perfectly parallel with the breadboard, which contributes to small errors in data generation.

For experiment 5 specifically, future work may include powering the mechanism with a motor with a set RPM instead of pushing it manually. Despite the fact that the filament holder is extremely smooth, human errors during pushing may contribute to potential errors in the signals received.

Also, the weight was not balanced on the wheel during the experiment. The MEGO portable power supply weighs 1 side of the wheel down significantly, contributing to a flawed sinusoid.

## 7. Conclusion

This study successfully demonstrated the utility of a Kalman-Filter infused algorithm that effectiveness to reduce measurement noise. The components of a low-cost MPU6050 sensor includes an accelerometer and a gyroscope, which both demonstrate their shortcomings. The accelerometer is prone to short-term fluctuations due to its high sensitivity towards noise and measurement errors; on the other hand, the gyroscope faces the challenge of cumulative error, which causes long-term drift.

The incorporation of a Kalman-Filter algorithm that combines the advantages of both signals was used in 5 separate experiments to test its performance under both static and kinetic conditions. Overall, the algorithm illustrates a much smoother tracking and was far less influenced by noise.

These results demonstrate the value of a Kalman filtering algorithm in real-world motion-tracking systems, where measurements are rarely perfect and multiple noisy data sources must be combined in order to piece a more holistic picture of the entire object tracking mechanism.

## Appendix

### *Raspberry Pi STEPico Pinout Map*

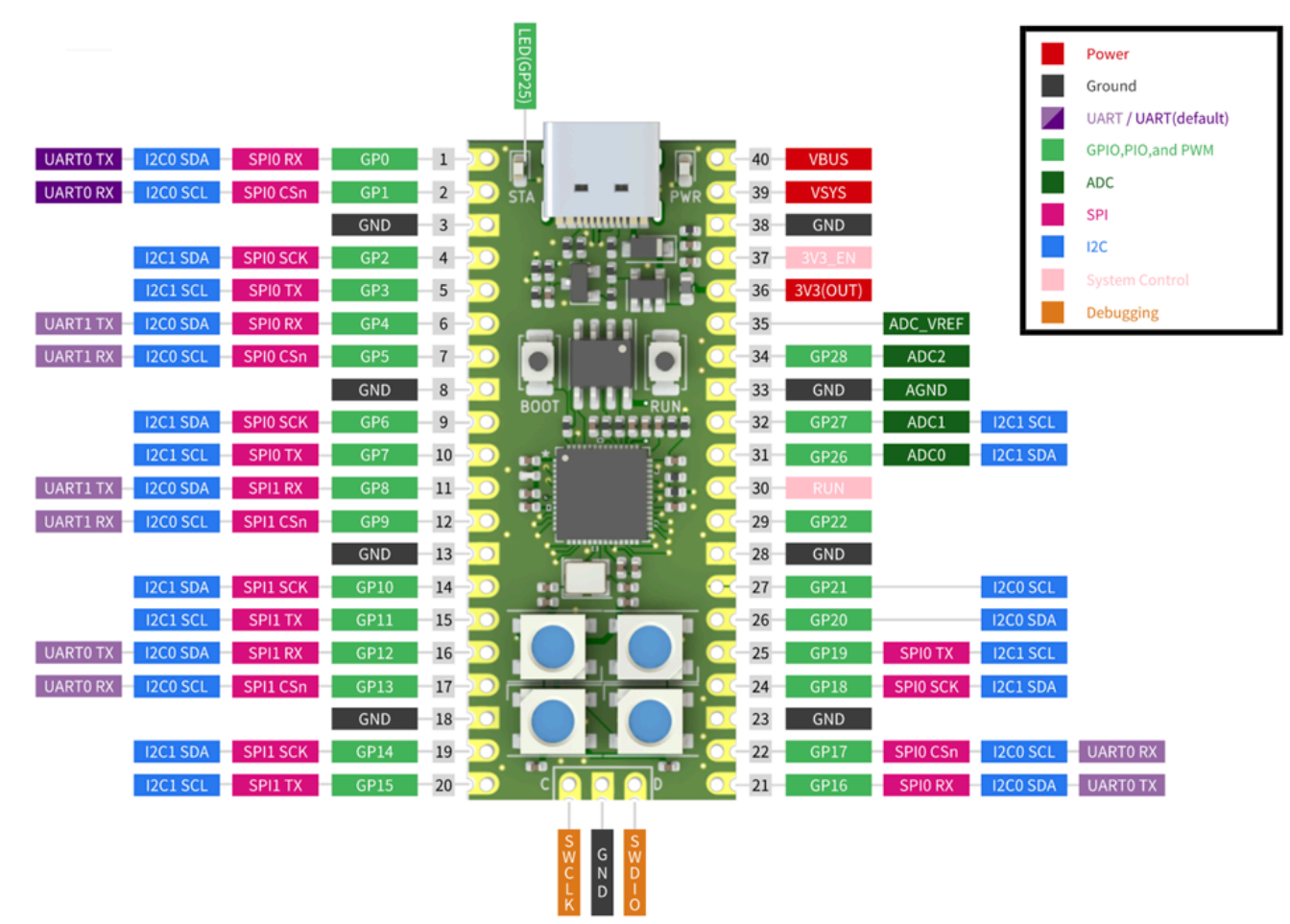

### *MPU6050 Pinout Map*

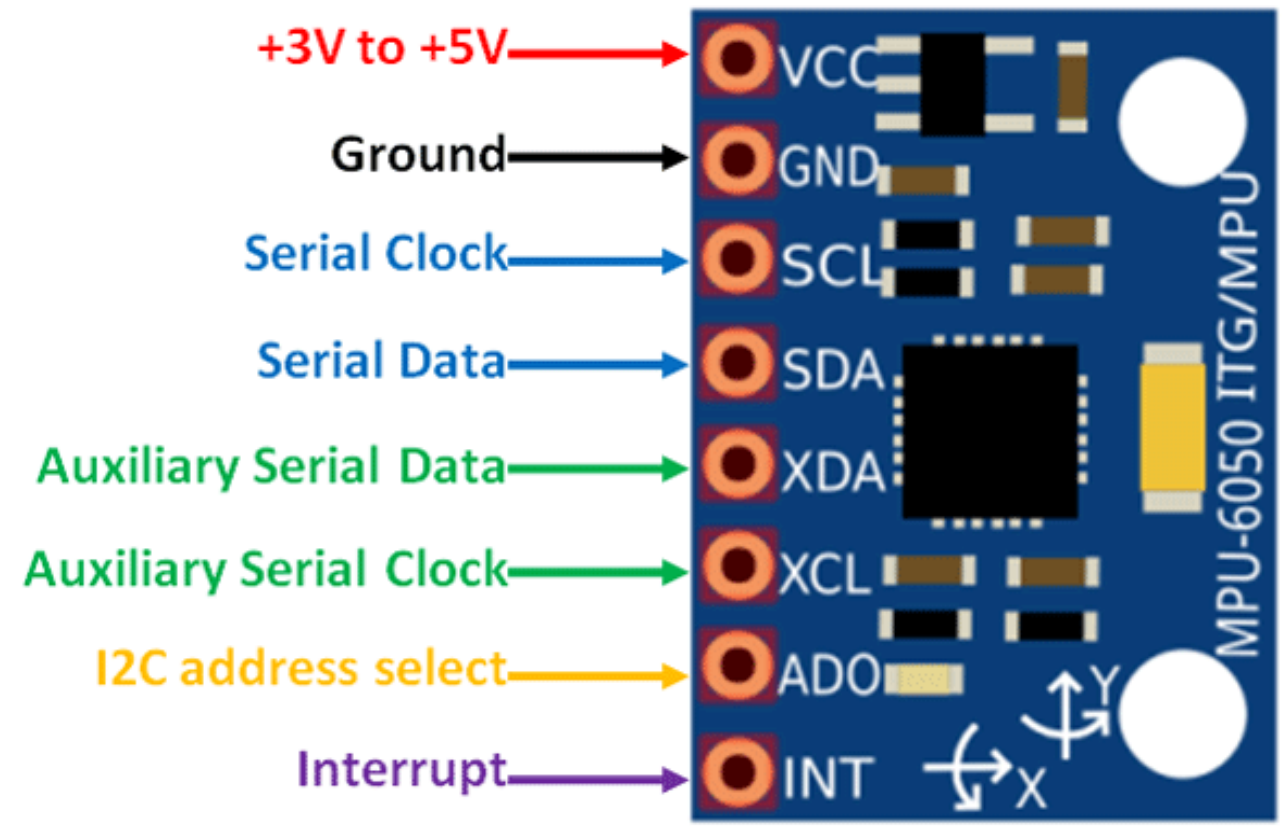

*Hardware Configuration Pin Mapping for Experiment 1-4*

| Function | RP2040 Pin | MPU6050 Pin |
|---|---|---|
| I2C SDA | GP | SDA |
| I2C SCL | GP | SCL |
| Power | Vsys | Vcc |
| Ground | Gnd | Gnd |

*Hardware Configuration Pin Mapping for Experiment 5*

| Function | RP2040 Pin | MPU6050 Pin |
|---|---|---|
| I2C SDA | GP | SDA |
| I2C SCL | GP | SCL |
| Power | Vsys | Vcc |
| Ground | Gnd | Gnd |
| Data Saving Indicator | GP18 | N/A |

*Software Infusion Guide*

*For installation guides and software codes (Thonny, R plots, etc), refer to GitHub Repository: https://github.com/KingofSaltyFish/Kalman-Filter-Research*